# Ambiguity-Driven Fuzzy C-Means Clustering: How to Detect Uncertain Clustered Records

Meysam Ghaffari
Department of Electrical and Computer Engineering,
Isfahan University of Technology
meysam.ghafari@ec.iut.ac.ir

Nasser Ghadiri[1]
Department of Electrical and Computer Engineering,
Isfahan University of Technology
nghadiri@cc.iut.ac.ir

*Abstract:* As a well-known clustering algorithm, Fuzzy C-Means (FCM) allows each input sample to belong to more than one cluster, providing more flexibility than non-fuzzy clustering methods. However, the accuracy of FCM is subject to false detections caused by noisy records, weak feature selection and low certainty of the algorithm in some cases. The false detections are very important in some decision-making application domains like network security and medical diagnosis, where weak decisions based on such false detections may lead to catastrophic outcomes. They are mainly emerged from making decisions about a subset of records that do not provide enough evidence to make a good decision. In this paper, we propose a method for detecting such ambiguous records in FCM by introducing a certainty factor to decrease invalid detections. This approach enables us to send the detected ambiguous records to another discrimination method for a deeper investigation, thus increasing the accuracy by lowering the error rate. Most of the records are still processed quickly and with low error rate which prevents performance loss compared to similar hybrid methods. Experimental results of applying the proposed method on several datasets from different domains show a significant decrease in error rate as well as improved sensitivity of the algorithm.

# 1   Introduction

Clustering is an unsupervised method for grouping data into several partitions. The key idea behind this method is that the algorithm tries to group similar data into one cluster, so the records in one cluster are more similar to each other and records in different clusters are more different from each other. The similarity of the records can be measured by computing the distance between each pair of the samples. Among many distance-based clustering algorithms, two widely used algorithms are K-means [1] and fuzzy c-means (FCM). Both algorithms are based on partitioning a database of $N$ objects into a set of $K$ clusters. These methods usually initialized with a single partition and then use an iterative control strategy to optimize an objective function and partition the space. One weaknesses of K-means is that it tries to put each sample in exactly one cluster, even if there is a lack of certainty and we are uncertain about selecting a single cluster to assign the sample to it. Therefore, some samples are falsely clustered and lead to weak decisions. On the other hand, FCM is based on the concept of fuzzy sets, allowing each sample to belong to more than one cluster at the same time, with possibly different degrees of membership to each cluster. Thus, it will assign each sample to several different clusters with possibly different degrees of membership, even if there is a lack of certainty [2]. FCM has its own shortcomings. For instance, initial choosing of cluster seeds will hugely affect the final result in some cases. To solve this problem, some improvements have been proposed for the FCM method, which help to select the initial seeds [3, 4]. Another problem with the FCM rises when the degrees of membership of a given sample in two or more clusters are almost equal. FCM does not perform as discriminating as expected when we have such *ambiguous* data in our samples. To the best of our knowledge, the concept of ambiguity of data in FCM clustering is not taken into account in previous methods. It will be emphasized throughout this paper that in many learning algorithms like FCM, there are situations that the algorithm is unable to infer the final results with the required high precision. In existing methods, the algorithm just tries to assign the sample to the most relevant cluster or even to a random cluster in such situations. In some applications, the effect of this weak decision making is too serious. There are many application domains where the precision of the algorithm is critical and thus we cannot bear high false rate. Examples are location privacy in mobile applications where people expect the system to prevent revealing their location, network intrusion detection where an attacker can compromise the system after a false detection, and medical diagnosis where critical life-threatening decisions have to be made based on false classification results.

[1] *Corresponding author. Address: Department of Electrical and Computer Engineering, Isfahan University of Technology, Isfahan, Iran. Phone : +98-311-391-9058, Fax: +98-311-391-2450, Alternate email: nghadiri@gmail.com*

The performance also matters in emerging clustering applications. The tremendous amount of information in many domains and the need to process more data with higher precision requires a method that preserves both accuracy and speed. This is still a real challenge in many application fields [2, 5]. Some improved methods are proposed that perform a trade-off between accuracy and performance [1, 3, 4, 6]. However, to meet the higher precision requirements of the aforementioned application domains, a considerable loss of speed will result prevents their efficient usage.

Our main contribution in this paper is to propose a method for improving FCM, namely Ambiguity-Driven FCM (AD-FCM) to tackle the aforementioned situations where both precision and performance are important. AD-FCM is based on the idea that if we analyze most of records with high precision, it would be possible to send the rest of them, the uncertain samples, to another learning or inference algorithm for gaining higher precision and higher confidence levels. This approach may lead to slightly longer execution time, compared to clustering or classifying all records with a lower precision. The proposed AD-FCM method is extensively evaluated by applying it to real data sets for intrusion detection, diagnosis of diabetes, image segmentation and location data privacy.

The rest of the paper is organized as follows. An overview of related work is given in Section 2. Section 3 describes the proposed AD-FCM method after a brief overview of basic FCM concepts. Experimental results with four real-world datasets in four different application domains are presented in Section 4. Section 5 concludes the paper and points out to future research.

## 2 Related work

Distance-based clustering algorithms are widely used as well-known family of machine learning methods. The first and most known clustering algorithm is K-means [1] which partitions the input space based on the distance between records and assigns each record to a single cluster. More flexibility in clustering was provided by introducing fuzzy clustering methods which allow each sample to belong to more than one cluster, with possibly different values of membership in each cluster. A widely used fuzzy clustering algorithm is the Fuzzy C-Means clustering algorithm [2]. In FCM, each cluster is treated like a fuzzy set and the cluster's centers move smoother than the K-Means during the execution of the clustering algorithm. Each sample will have its own effect on the center of each cluster with some fuzzy factor, unlike K-Means which allows each sample to be a member of just a single cluster. However, a problem with the FCM algorithm is the initial selection of cluster seeds which may affect the final results of partitioning the input space, leading to weak or unstable decisions which is a real challenge in some application areas such as network security and medical diagnosis. Many improvements have been developed for the basic FCM algorithm. Yang proposed penalized FCM [5] which tries to maximize the similarity of clusters. Krishnapuram and Keller proposed a probabilistic version of FCM to overcome the initial seeds effect [7]. Siraj and Vaughn [8] use Fuzzy Cognitive Maps [9] to create the initial clusters for preventing high impact on the final results. For a better discrimination between clusters, Höppner and Klawonn [10] offered an improved method for detecting the partition boundaries in FCM. Another interesting contribution was collaborative FCM proposed by Pedrycz and Rai [11]. In their method, the clusters interact with each other to improve the final results. Kernel-based and multiple kernel methods [12] were also two other improvements on FCM. In a recent work, Mei and Chen [13] proposed relation integrated FCM which also performs the fuzzy clustering in a cooperative manner and uses pairwise analysis of data samples and the relationships between them.

Most of the improvements on FCM have been focused on improving the accuracy of the algorithm and decreasing the clustering errors. However, to the best of our knowledge, none of them directly addresses the problem of ambiguity of the similar cluster membership values, which leads to lack of certainty in the FCM clustering output. Even though the FCM is already used in some specific application domains like intrusion detection [14] this shortcoming of FCM has not been taken into account. We argue that uncertainty is an inherent and non-separable property of most machine learning algorithms and more specifically in the distance-based clustering algorithms. Our proposed AD-FCM method addresses this problem to prevent false decisions when we are uncertain about the cluster memberships, leading to more accurate results.

Recent studies also confirm the general lack of accuracy in learning algorithms [15]. It has been argued that there is no perfect learning technique [16]. This has led to exploiting hybrid techniques, which combine the results of different machine learning techniques for improving the accuracy [17]. In general, hybrid machine learning methods have better results than a single method [18, 19]. However, a big challenge for hybrid methods in many emerging application domains is the computational complexity. For instance, in network intrusion detection, the problem is the tremendous growth of infrastructure and higher bandwidth which requires efficient processing [16]. The hybrid learning techniques rely on heavy computations and their response time is affected by the processing speed of the specific machine learning algorithm, where the faster techniques are clearly preferable [18, 20]. Although hybrid techniques provide better accuracy, but they execute several machine learning methods over the whole dataset and then merge the results. Therefore, their processing time will increase significantly. This is an important barrier which prevents using the hybrid methods in real-world applications. In contrast, our proposed AD-FCM method provides accurate results similar to that of hybrid techniques, without the need to run multiple algorithms over the whole dataset. Instead, it will identify the ambiguous records, then it separates and sends only these ambiguous records to another algorithm for further processing. So instead of sending the whole dataset, only few number of records will be processed multiple times. This allows the AD-FCM to improve the accuracy with negligible processing overhead.

## 3 Ambiguity-Driven FCM clustering

This section begins with an overview of the basic FCM clustering method. The problem of ambiguity in assigning a sample to the FCM clusters is discussed in the next part, followed by describing the proposed AD-FCM process for coping with this ambiguity. The symbols and notations used throughout the paper are summarized in Table 1.

**Table 1.** Symbols and notations

| Symbol | Description |
|---|---|
| *P(h) = P(hypothesis)= P(C) = P(cluster)* | The probability value of a hypothesis. Here, our hypothesis is belonging to a specific cluster where the probability is equal to the average of membership values of the records to the specific cluster for the dominant cluster (membership>=0.5) and the average of fuzzy complements of the membership vaules for non-dominant clusters. |
| *P(S/h), P(S/C)* | The value of membership of a record in a specific cluster for dominant cluster (membership>=0.5) and the fuzzy complement of the membership value for non-dominant clusters. |
| *C* | Number of clusters |
| **Certainty Threshold** | A threshold to distinguish between certain and ambiguous records |
| **PAR** | Percent of Ambiguous Records |
| **PBFRA** | Percent of Basic False detected Records within Ambiguous records (percent of the records falsely clustered by FCM and separated by AD-FCM as the ambiguous record within all separated ambiguous records) |
| **NAR** | Number of Ambiguous Records |
| **NTR** | Number of True Records |
| **NFR** | Number of False Records |
| **NFRA** | Number of False Records within Ambiguous records |

## 3.1 The basic FCM method

As mentioned in Section 1, distance-based clustering is an unsupervised method for grouping a set of data records into different clusters. The classic K-Means clustering assigns each record to exactly one cluster. By contrast, FCM clustering uses fuzzy sets to partition a database of *N* objects into a set of *K* clusters in such a way that it allows each sample to belong to more than one cluster, with possibly different degrees of membership in each cluster. Thus, it clusters the input with different degrees of membership even if there is a lack of certainty. The main idea of the FCM algorithm is based on getting a collection of n-dimensional records as input, and trying to minimize the objective function shown in Eq.(1) to assign the records to clusters [2]:

$$Q = \sum_{i=1}^{c} \sum_{k=1}^{N} u_{ik}^{m} ||X_k - V_i||^2 \qquad (1)$$

Here $V_i'$ is an n-dimensional prototype for each cluster, and *u* is a partition matrix that assigns the degrees of membership to each sample. The coefficient *m* is called fuzzification coefficient which determines the degree of fuzziness of memberships on clusters. By increasing the value of *m*, the fuzziness of the algorithm will be increased. The objective function of FCM clustering is also shown by:

$$Q = \sum_{i=1}^{c} \sum_{k=1}^{N} u_{ik}^{m} \sum_{j=1}^{n} (x_{kj} - v_{ij})^2 \qquad (2)$$

where $x_{kj}$ is the input record, $v_{ij}$ is center of the cluster and *u* is the objective function, *N* shows the number of input records and *c* is number of clusters.

## 3.2 Ambiguity in clustering algorithms

Most clustering algorithms try to partition the records into a set of specific clusters. In some cases, it is very difficult to determine the real clusters with the given set of attributes of samples or with specific algorithms. Consider two adjacent crisp clusters in two-dimensional space with one shared border. What happens if we have a sample on this border? Using current methods, the algorithm tries to assign that sample(s) either to the most relevant clusters in fuzzy clustering or to a random cluster in non-fuzzy clustering. The fuzzy case is illustrated in Figure 1. Two fuzzy clusters exist and the space between them is where records cannot be assigned to any specific cluster with enough certainty, i.e. as certain as the core of clusters. We name the space between clusters the "ambiguous space". It can be observed that when a sample perches here, the exact cluster which this sample belongs to, cannot be determined with the same confidence level of other areas of the cluster, such as areas close to its core. This method of making decision about the clusters that the record belong to , is not perfect enough in most cases because of its lack of accuracy which leads to significant increase in error rate. Therefore, we try to propose a method for determining most of ambiguous records, which makes it possible to increase the accuracy.

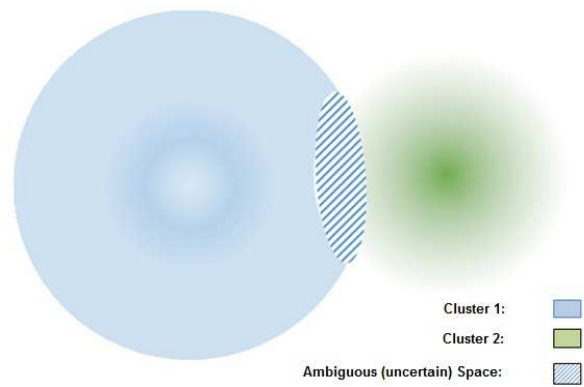

**Figure 1:** The ambiguous space in clustering

In some application domains like intrusion detection or medical diagnosis, a very important aspect is the confidence of the results. So if some ambiguous records exist when performing FCM, they can be detected using our method and investigated with other methods. This will potentially provide better accuracy. Discriminating the ambiguous samples out of the unambiguous ones will cause the final results to be more reliable than existing clustering methods.

## 3.3 The proposed AD-FCM method

As discussed in Section 1, one of the biggest challenges in clustering is the lack of certainty in discriminating between ambiguous and unambiguous samples. In this section, we present the design of the AD-FCM, which aims at coping with the ambiguity that exists in fuzzy clustering. AD-FCM is designed to provide more accuracy than existing methods, but, unlike the hybrid techniques, with little or no negative impact on processing time. This is achieved by a two-step clustering approach. In the first step, most of the data samples are clustered with high accuracy and high speed. In the second step, the ambiguous records will be sent for further processing. By this design, the benefits of hybrid algorithms will be preserved, but their extensive time consumption is avoided. The proposed algorithm for clustering with detection of ambiguous records is described below.

The input records with selected features are sent to FCM for clustering. After finishing the FCM algorithm, the results have to be defuzzified and assigned to their maximum relevant cluster. In this step, instead of assigning the input record to the cluster with higher membership value –as commonly done in classic FCM–, we use a Bayesian method to aggregate all clusters' decisions about each record. Let's have a brief overview of the Bayes theorem first, and customize it for our purpose in the next part

### 3.3.1 The Bayes' theorem and certainty factor

Given an input record $R$, it is useful to determine the certainty of a given hypothesis or the best hypothesis about $R$ drawn from space $H$. One way is calculating the probability of the hypothesis with the given record $R$ and the initial knowledge. The Bayes' theorem provides a method to calculate such probabilities. In other words, the Bayes' theorem makes it possible to calculate the probability of the hypothesis based on its prior probability, the probabilities of other different records and different hypotheses, and the observed record itself. Let's introduce some notations for a better description of the concept.

We assume that $P(h)$ is the initial probability that hypothesis $h$ holds. For a given cluster $C$, $P(h)$ is calculated as the average of the membership values of all records in that cluster. In addition, for calculating the validity of this value based on membership of $R$ in other clusters except $C$, we require other clusters to admit the degree of dependency of the records to them by *complementing* the membership values of $R$ in those clusters. The idea behind the complement operation is that it shows the probability of a record *not* belonging to other clusters. In other words, it will show the aggregated probability of not happening $h$, called $h'$, where $h'$ consists of all hypotheses except $h$. We also define $P(s|C_i)$ as the score of a given record in the expected hypothesis which is extracted as below:

$P(s|C_i)$ is calculated as the membership value of a given record in the relevant cluster $C_i$ or in the expected cluster. For irrelevant clusters or non-expected hypothesis $P(s|C_i)$ is calculated as the complement of all membership values.

$P(C_i/R)$ is probability of cluster $i$ with given record $R$. Now the basic Bayes' theorem can be defined by.

$$P(C_i|R) = \frac{P(s|C_i)P(C_i)}{P(R)} \qquad (3)$$

For any given record *R*, if we assume *P(R)* equal to 1 for the probability of membership of a given record in the expected cluster as resulted from FCM clustering, then we can introduce the certainty factor as:

$$Certainty\ Factor = \frac{\sum_{i=1}^{c} P(s_i|C_i)P(C_i)}{c} \qquad (4)$$

As mentioned earlier, $P(s_i|C_i)$ is the membership value of a given record in the relevant cluster or the complement of membership value in all irrelevant clusters, and *c* is the total number of clusters.

### 3.3.2 The general process of AD-FCM

An overall process of our proposed AD-FCM method is depicted in Fig. 2. It begins with a base FCM clustering and then performs the additional processing required to detect the ambiguous records. A pseudo code is also given to declare the detailed steps for calculating the certainty factor as presented in Fig. 3.

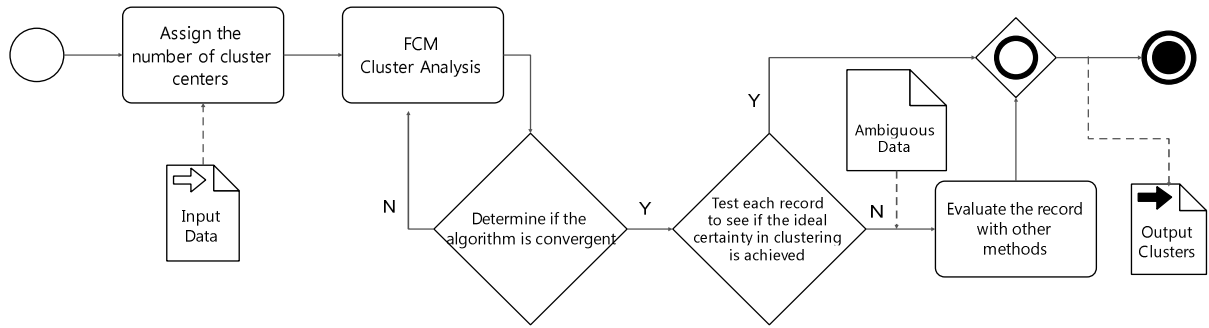

**Figure 2:** The AD-FCM process

The proposed AD-FCM algorithm begins with giving the *certainty threshold* that will determine the border between the unambiguous records (accepted by the clusters) and the ambiguous records. At line 2, the input data will be clustered by the FCM and then at lines 3 to 6 the matrix *P* which is the probability matrix will be calculated. Line 7 begins the loop for each records to calculate the maximum relevant cluster at line 8. Then in lines 10 to 13 the score is calculated. The computation of the formula as depicted in Eq. (4) is calculated in lines 14 and 15.

Finally, the input record will be labeled in lines 16 to 19 as ambiguous or member of the most relevant cluster based on comparing certainty factor with the certainty threshold. To explain how this rule works, note that it specifies the certainty of the record being a member of the higher related cluster and if its value is more than the threshold, called *certainty threshold* here, then we can assign the input record to that cluster with certainty. If the certainty factor is beyond the threshold, it means that the record has a tendency to other clusters either, and we need to check this record more carefully and label this record as *ambiguous*.

```
1. Get Certainty threshold
2. Cluster data by FCM
3. For each i in clusters
4.      For k=1 to C
5.           If k=i, P(i,k)=average membership functions for cluster i
6.           Else P(i,k)= average complement membership functions for cluster i
7. For each record such as $R_j(R_1,..,R_j,..R_n)$
8.      $C' = argmax_j(U[j][i])$
9.      For each cluster $C_k$
10.          If $C_k = C'$
11.               $S_k = U[C_k][i]$
12.          Else
13.               $S_k = 1 - U[C_k][i]$
14.          $\Sigma = \Sigma + S_k * P[i][k]$
15.     Certainty factor=$\frac{\Sigma}{c}$
16.     If Certainty factor lower than Certainty threshold
17.          The record is ambiguous
18.     Else
19.          The record belongs to the C'
```

**Figure 3.** The pseudo code of AD-FCM

We can illustrate the algorithm with a simple sample. Suppose that we have four records and three clusters and degrees of membership of records in clusters are as shown in Table 2:

**Table 2.** The membership values of records in clusters

|  | Record 1 | Record 2 | Record 3 | Record 4 |
|---|---|---|---|---|
| **Cluster 1** | 0.1 | 0.8 | 0 | 0.1 |
| **Cluster 2** | 0.6 | 0.1 | 0.2 | 0.9 |
| **Cluster 3** | 0.3 | 0.1 | 0.8 | 0 |

Then matrix *P* is calculated as below:

*P(1,1)*: since $i = j$ then we have *P(1,1)*= average (membership values of cluster 1)

*P(1,2)*: sinceas $i\ !=\ j$ then we have *P(1,2)*= average (complement membership values of cluster 2)

Thus we have:

$$P := \begin{bmatrix} 0.25 & 0.55 & 0.7 \\ 0.75 & 0.45 & 0.7 \\ 0.75 & 0.55 & 0.3 \end{bmatrix}$$

Based on matrix *P*, the certainty factor of records is calculated as follows. We have to determine the most relevant cluster (maximum argument of the *C*) so that the specific row of matrix P can be extracted and the scores can be calculated based on this relevant cluster, thus certainty factor is calculated as:

$$Certainty\ factor\ for\ R_1 = \frac{(0.9 * 0.75) + (0.6 * 0.45) + (0.7 * 0.7)}{3} = 0.4784$$

$$Certainty\ factor\ for\ R_4 = \frac{(0.75 * 0.9) + (0.45 * 0.9) + (1 * 0.7)}{3} = 0.5934$$

# 4 Experimental Results

In this section, the proposed AD-FCM clustering method is evaluated using four data sets. The well-known KDD99 data set for network intrusion detection systems (IDS), Pima Indian's diabetes as a medical dataset, the benchmark for semantic image segmentation and finally the MSR GPS Privacy mobile users' location dataset to evaluate effects of noise on basic FCM and AD-FCM. The similarity of the first two datasets is that both have a major normal class and one or more abnormal classes which are minor and sometimes unpredictable, thus hard to be learnt by learning algorithms. For intrusion detection, the goal is to differentiate intrusions from normal traffic. The general approach is to detect abnormal traffic which may be unseen before. Learning methods cannot be applied because of their processing times and their inability to detect unseen patterns. For the Pima data set, there are imbalanced classes that may help the learning algorithms to provide higher precision for the major class, but minor classes are detected with low precision making these algorithms unreliable. In image segmentation data set, it is shown that AD-FCM determines the ambiguous parts that falsely are clustered in the basic algorithm and thus can increase the accuracy of applying FCM for image segmentation. Finally, the MSR dataset where AD-FCM is applied to potentially provide more robustness to noises that are added to the dataset to achieve privacy. Thus, the AD-FCM is expected to provide more reliable results than existing methods in different application areas.

## 4.1 AD-FCM for intrusion detection

One of the biggest challenges in IDSs is the trade-off between the computation and keeping the systems secure. In most of the IDS scenarios the packets cannot be investigated deeply. Nowadays algorithms investigate the packets almost careless and superficial in order to relieve heavy computations [1, 3, 4, 6] and this is an unsolved problem [2, 5]. The proposed AD-FCM method is applied to IDS data to demonstrate that unlike existing methods, it can provide both benefits, speed that usually comes with simple learning methods and high accuracy that requires hybrid methods.

### 4.1.1 Data preprocessing and feature selection

An important step in traffic classification is feature selection. A small and optimal subset of features is generally required, or the learning algorithm will be misled or overloaded. Many approaches exist for IDS feature selection [21]. We use a modified Breadth First Search (BFS) method, which works as described below.

For feature selection, we discretized features in WEKA [22]. Then we define *H(x)* as the information entropy for variables such as *X*. *X* can take $N_X$ discrete values. The relative uncertainty (RU) [23] for variable *X* is shown as Eq.(5) below:

$$RU(X) = \frac{H(X)}{H_{max}(X)} = \frac{H(X)}{log_2(min/N_X, m/)} \qquad (5)$$

The conditional RU is given in Eq.(4) below ehre $C_j$ is the *jth* class and $A_i$ may take $N_{Ai}$ number of values. $N_{jik}$ shows the number of samples whose values of *Ai* is equal to *kth* value in *Ai* which belongs to $C_j$.

$$RU(A_i/C_j) = \frac{\sum_{k=1}^{N_{Ai}}(-p(A_{ik}/C_j)*log_2(p(A_{ik}/C_j)))}{log_2(min/N_{Cj}, N_{Ai}/)} \qquad (6)$$

and

$$p(A_{ik}/C_j) = N_{jik}/N_{Cj}$$

$A_i$'s bias coefficient is defined as:

$$B(A_i/C_j) = 1 - RU(A_i/C_j)$$

So if the bias coefficient of an attribute is higher, it will be assigned to $C_j$ with more certainty than other clusters. Then symmetric uncertainty (SU) is used to measure the discriminating ability of features. The *SU* between $A_i$ and *C* is defined as:

$$SU(A_i, C) = 2[\frac{IG(A_i|C)}{H(A_i)+H(C)}] \qquad (7)$$

SU is calculated for each pair of a feature and a class, which gives an array of *SUs*, and the feature subsets are selected based on this list [24].

### 4.1.2 Clustering the IDS Data

In this step, the input records with selected features are sent to AD-FCM. As described in Section 3.3, the AD-FCM has two types of output: known data and ambiguous data. Table 3 shows the best results of the AD-FCM which depends on the number of clusters and the fuzzification coefficient (*m*) which identify the fuzziness of the algorithm and depicted in Eq. (2) as the exponent of the *U* where *U* is the objective function, since the proposed AD-FCM inherits the general properties of the FCM. As there are more than one type of abnormal traffic, the precise number of clusters based on data has led to better results. Best results are achieved with 5 clusters and the best value of fuzzification coefficient was *m*=2 as shown in Table 3.

**Table 3:** The best results of running AD-FCM with different number of clusters and different fuzzification coefficients (*m*)

| Number of clusters | *m* | False detection rate |
|---|---|---|
| 3 | 2 | 1% |
| 3 | 3.2 | 1.5% |
| 4 | 2 | 2% |
| 4 | 3.2 | 1% |
| 5 | 2 | 0.7% |
| 5 | 3.2 | 2% |

The results of increasing the certainty threshold are shown in Table 4 and Table 5. An interesting observation is that when the number of ambiguous records is increased, the percentage of basic false detected records –records that are clustered falsely in basic FCM among all determined records as ambiguous in AD-FCM- will decrease. A balance is required here, since more ambiguous records in the dataset will decrease the detection error. However, having more ambiguous records means that a deeper investigation is required for a higher number of records, which leads to increased requirement to computational resources. So the trade-off between handling ambiguity and performance will guide the selection of the certainty threshold value.

**Table 4:** The effect of the number of ambiguous records on the ratio of basic falsely detected records among ambiguous records

| Clusters | *m* | Certainty | All ambiguous data | False - basic detection |
|---|---|---|---|---|
| 5 | 2 | 0.45 | 5.9% | 37% |
| 5 | 2 | 0.25 | 1% | 53% |

**Table 5:** The effect of changing *certainty threshold* on ambiguous data and false records

| Ambiguous | Certainty Threshold | False Best | False Mean | False Variance (10 test) |
|---|---|---|---|---|
| 1% | 0.25 | 3.1% | 3.66% | **0.1214** |
| 2.8% | 0.3 | 1.9% | 1.74% | **0.2854** |
| 3.1% | 0.35 | 1% | 1.63% | **0.0368** |
| 3.7% | 0.4 | 0.75% | 0.79% | **0.0059** |
| 5.9% | 0.45 | 0.7% | 0.86% | **0.019** |

Fig. 4 shows the effect of changing the values of certainty threshold. By increasing this factor, the amount of ambiguous records will be increased and the number of false records will be decreased. As depicted in Figure 4, the ambiguous records are increased significantly above the threshold of 0.4, but basic false detections do not decrease that much. It can be interpreted as if we increase the sensitivity of the AD-FCM after the optimum point, we are just increasing the computation overhead because we need to reprocess these records without separating many false records. The optimum value of the certainty threshold in this experiment was 0.4. Determining the optimum value seems to be application dependent but it can be easily computed by a few experiments. Our experiments with other datasets also show that the optimum range is between 0.3 and 0.4.

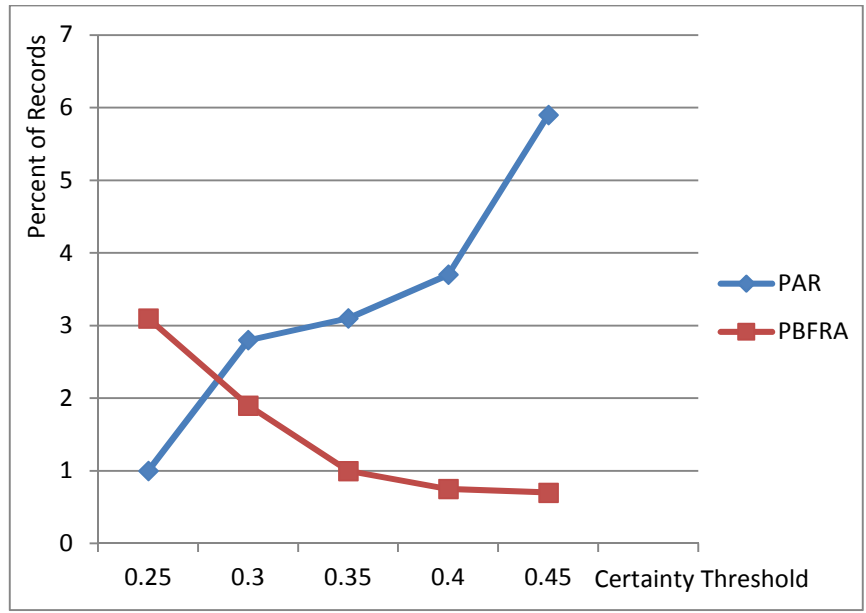

**Figure 4:** The effect of increasing *Certainty threshold* on percent of ambiguous records (from all records) and percent of basic false detected records (from ambiguous records)

Table 6 summarizes the comparison of the proposed AD-FCM method with some of well-known and relatively fast methods. The first compared method is a modified version of Support Vector Machine (SVM) which offers an accuracy of 97.17% in specific cases with a minimum of 6% of falsely detected records for bulk network traffic [25]. It was slower than AD-FCM method. Other methods were K-star, J48, and BFTree [26], K-means[27], and modified C4.5 which uses extended data for classification [28].

**Table 6:** Comparing the proposed AD-FCM method with other methods

| Method | False Detection Rate |
|---|---|
| SVM-based Method | 6% (3% for specific data) |
| K-star | 5.97% |
| BFTree | 10.65% |
| J48 | 13.01% |
| K-means | 3.86% |
| C4.5 with extended data | 3.3% |
| AD-FCM | <1% |

The overall interpretation of the results of the AD-FCM for IDS shows that instead of false clustering of some records, we have an *unknown* class that contains the data records that are ambiguous for the algorithm. This helps us to cluster data in two steps: most of the input data are clustered very fast in the first step as usual, and in step two we analyze the rest of the dataset more carefully using another method. This general approach provides better results than FCM or hybrid methods, since most of the existing algorithms have low accuracy or high computational cost, and we need to choose between them. In contrast, with AD-FCM method we can take the advantage of both.

## 4.2  AD-FCM for Medical Diagnosis

As a second evaluation of the AD-FCM, in this section we use Pima Indian diabetes data from the UCI Repository of machine learning database. The dataset consists of 768 records with 8 attributes. We use two clusters: one for the normal records and another for sickness. The results of applying AD-FCM are shown in Table 7. It can be observed that the false detection rate decreases with increasing certainty threshold. The results of these experiments are also depicted in Figure 5.

**Table 7:** The effect of certainty threshold on false detection rate ($m=2$)

| Threshold | False No | Ambiguous No | Percent of ambiguous records, invalid detection in basic mode |
|---|---|---|---|
| **0** | 264 | 0 | 0 |
| **0.15** | 261 | 6 | 50 |
| **0.2** | 237 | 50 | 54 |
| **0.25** | 202 | 122 | 50.81 |
| **0.3** | 166 | 198 | 49.49 |
| **0.35** | 166 | 198 | 49.49 |
| **0.4** | 164 | 207 | 48.31 |
| **0.45** | 158 | 223 | 47.53 |

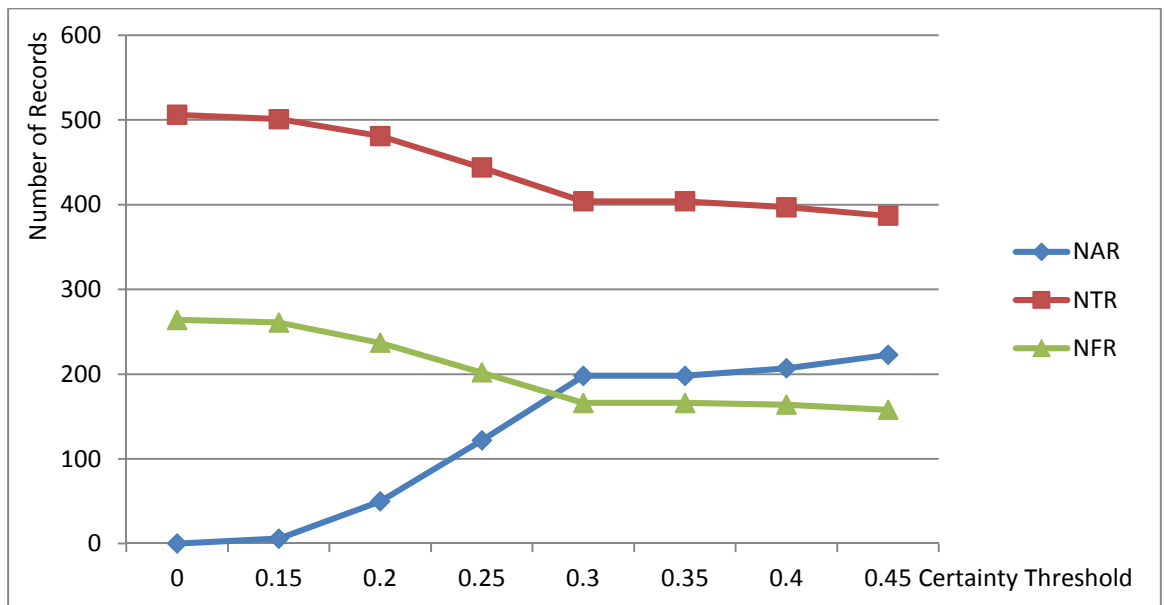

**Figure 5:** Ambiguous data vs. false data (number of ambiguous records and false records and truly clustered records in output of the AD-FCM)

As depicted in Figure 6, by increasing the certainty threshold, the number of ambiguous records will increase and the ratio of false clustered records will decrease. It can be also observed on Figure 6 that the best rate of basic falsely detection occurs at certainty threshold is in range of [0.3-0.4], as discussed earlier in this section.

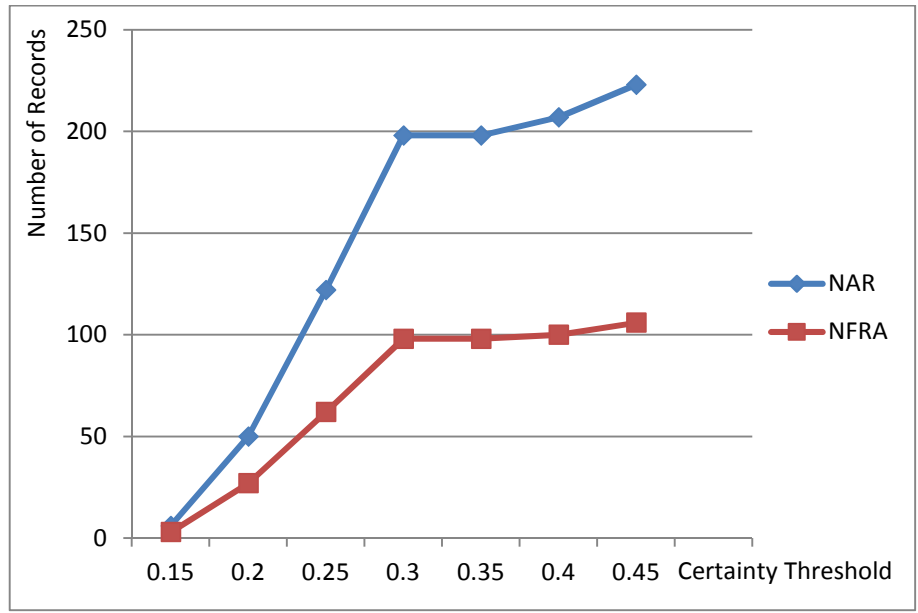

**Figure 6:** Number of false detected records using basic method vs. increasing ambiguous data by increasing certainty threshold

Table 8 shows the effect of increasing certainty threshold on the accuracy of the output. Note that the accuracy is calculated as below:

$$Accuracy = \frac{True\ Clustered\ Samples}{True\ Clustered\ Samples + False\ Clustered\ Samples} \qquad (8)$$

We used this equation instead of the traditional one since we need to separate ambiguous records and we cannot count them here. The table shows that by increasing certainty threshold, the accuracy of output results will be increased. Also we can observe in the same type of dataset, if the basic method has better results, the AD-FCM prospers the final results more efficiently. This is due to the fact that as we use each cluster as a hypothesis and we infer based on these hypotheses, the more accurate hypothesis could have better results. These results in Table 8 show that for a different dataset like Pima, where the minor class is smaller and FCM has better results, the effect of increasing certainty threshold on the accuracy of AD-FCM is more significant.

**Table 8:** The Accuracy of the AD-FCM with different certainty threshold in case of having different distribution of classes in dataset

| Certainty Threshold | Complete dataset | Minor class is 35% | Minor class is 10% | Minor class is 5% |
|---|---|---|---|---|
| 0 | 65.88% | 70.51% | 79.82% | 80.57% |
| 0.15 | 65.74% | 71.3% | 81.15% | 81.66% |
| 0.2 | 66.99% | 73% | 84.52% | 85.83% |
| 0.25 | 66.66% | 76.84% | 92.71% | 96.64% |
| 0.3 | 70.87% | 79.22% | 96.19% | 98.53% |
| 0.35 | 70.87% | 79.22% | 96.19% | 98.53% |
| 0.4 | 70.76% | 79.84% | 96.17% | 98.51% |
| 0.45 | 71.01% | 79.72% | 96.09% | 98.49% |

Table 9 compares the results of the proposed AD-FCM method with C4.5, KAIG [29], SVM, and modified SVM for imbalanced classes [30]. It can be observed that in imbalanced datasets which have a major class and one or more minor classes, as the AD-FCM method is derived from FCM, it works well and has no tendency to the major class or cluster. The G-mean values show that when the minor class becomes smaller, other classification methods will tend to the major class. Note that as the FCM does not consider the size of clusters, unlike Bayesian theory, the best results for G-mean is achieved by lower certainty threshold where modified Bayesian theory has a lower effect. Best results for accuracy is achieved with certainty threshold 0.4 where the effect of modified Bayesian theory is more than the first case, and for G-mean we use a certainty threshold of 0.15. By increasing the certainty threshold (as the ambiguous records especially in minor cluster increase), the G-mean decreases because the modified Bayesian method works similar to other methods and it will detect more records in the smaller cluster as ambiguous. In other words, by increasing the certainty threshold, the accuracy of output data will be increased. The results of changing the certainty threshold on the accuracy of output are shown in Table 9.

**Table 9:** Comparing AD-FCM with modified SVM, SVM, C4.5 and KAIG

| Parameter | AD-FCM mean | Modified SVM for Imbalance classes Mean | SVM mean | C4.5 mean | KAIG mean |
|---|---|---|---|---|---|
| **35%** | | | | | |
| Overall acc. | 79.84 | 77.82 | 75.52 | 74.22 | 78.78 |
| G mean | 60.25 | 77.72 | 63.56 | 68.99 | 76.46 |
| **10%** | | | | | |
| Overall acc. | 96.17 | 87.64 | 89.93 | 88.49 | 87.05 |
| G mean | 83.38 | 71.5 | 0 | 23.63 | 59.73 |
| **5%** | | | | | |
| Overall acc. | 98.51 | 96.49 | 94.7 | 93.56 | 94.89 |
| G mean | 87.6 | 67.11 | 0 | 23 | 44 |

These comparison results are also illustrated in Figure 7. One can observe that by changing the balance of classes, AD-FCM method works more robust than other methods.

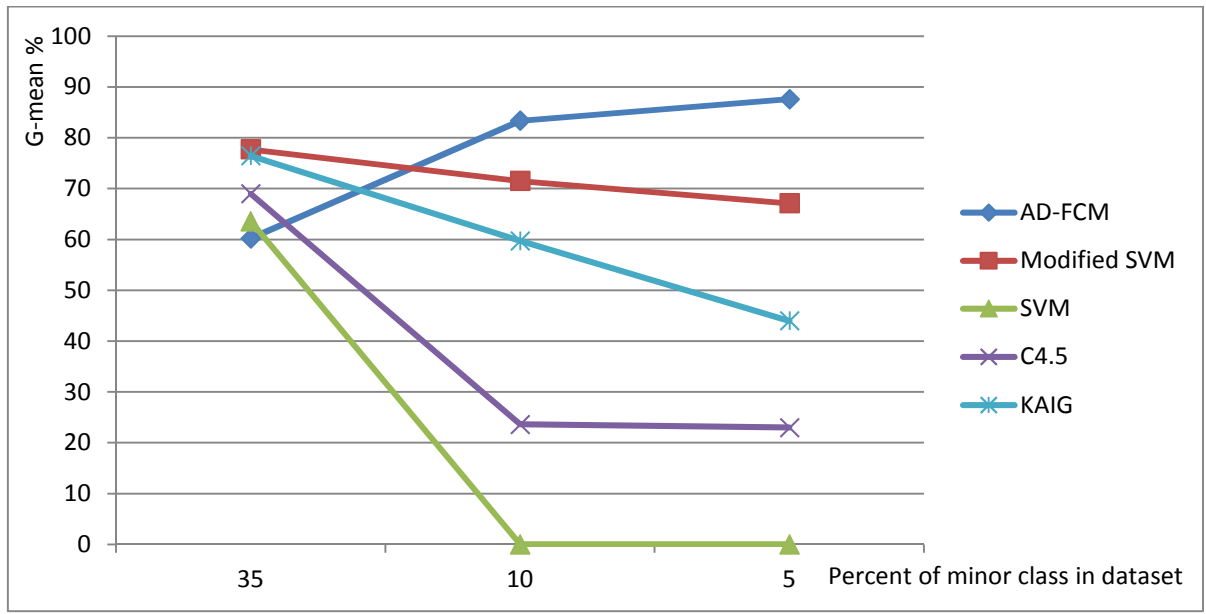

**Figure 7:** Comparing G-mean value of different methods for 3 imbalanced classes (when minor class is 35,10 and 5 percent of all dataset) in the Pima dataset

### 4.3 AD-FCM for image segmentation

One of the demanding application areas of FCM is image segmentation [31]. Here we compare FCM and AD-FCM for this purpose, and we analyze the effect of certainty threshold on the output. For simplicity we have converted the images into grayscale. The ambiguous parts in images are always shown by pure black color. The photos are taken from the benchmark for semantic image segmentation [32]. We also added a high quality photo for better illustration of the results.

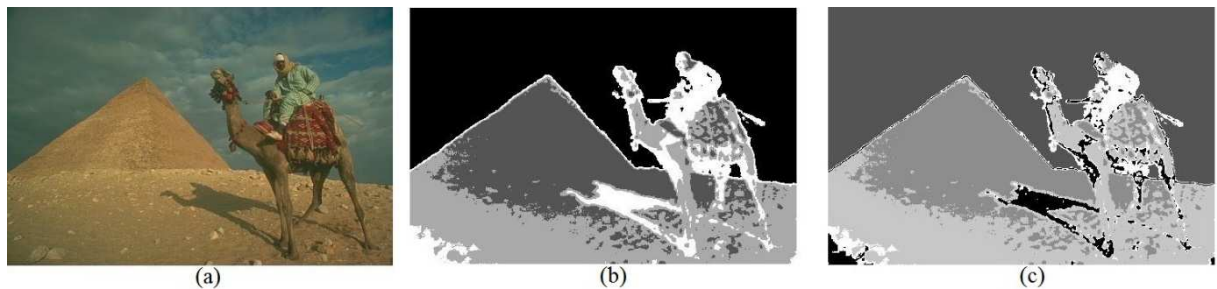

**Figure 8.** Image segmentation by FCM and AD-FCM: (a) the original photo, (b) segmented using FCM, (c) segmented using AD-FCM with certainty threshold of 0.4

As depicted in Fig. 8(c), the shadow of the camel is distinguished as ambiguous, which was almost the same in basic FCM. Fig. 9 also shows the effect of certainty threshold for another classic photo.

In Fig. 10(b) which shows the basic FCM, too many details are discarded, unlike Fig. 10(c) which shows AD-FCM result with a certainty threshold of 0.4 and the weeds are well detected by determining ambiguous parts. In Fig. 10(d) with a certainty threshold of 0.5 the face of the rhino is well analyzed. The animal's ear is better separated and inside of its ear is determined as ambiguous and is not in the same cluster as the other parts of its ear which is distinguishable by human in original image either. With increasing the certainty threshold in Fig. 10(e) most of the picture is marked as ambiguous.

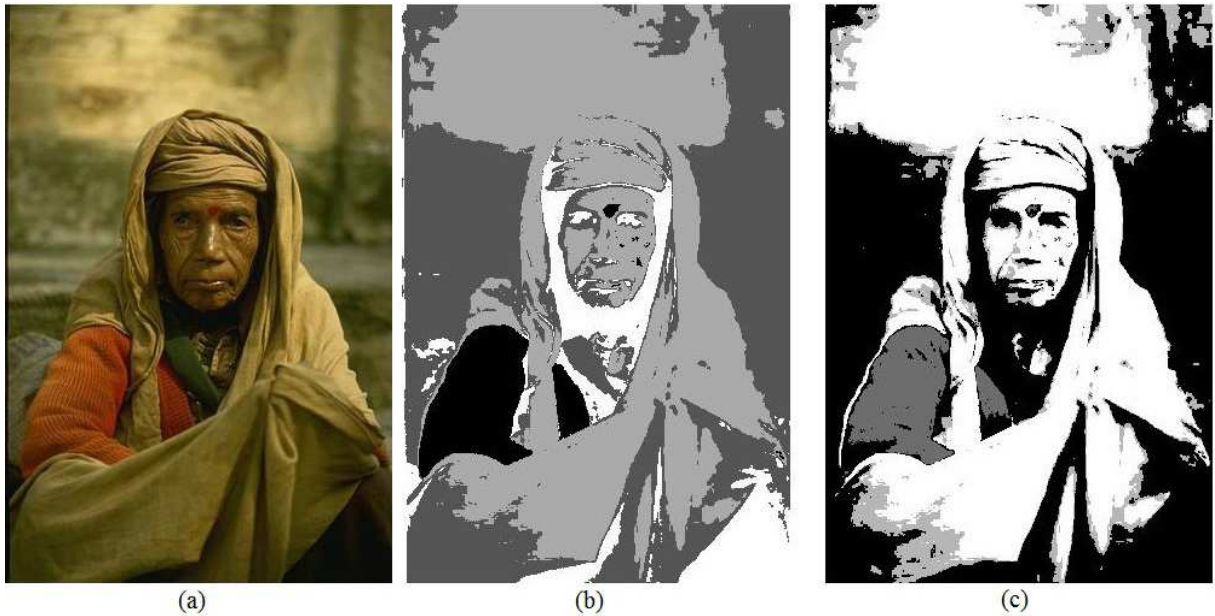

**Figure 9.** Image segmentation by FCM and AD-FCM: (a) the original photo, (b) segmented using FCM, (c) segmented using AD-FCM with certainty threshold of 0.5

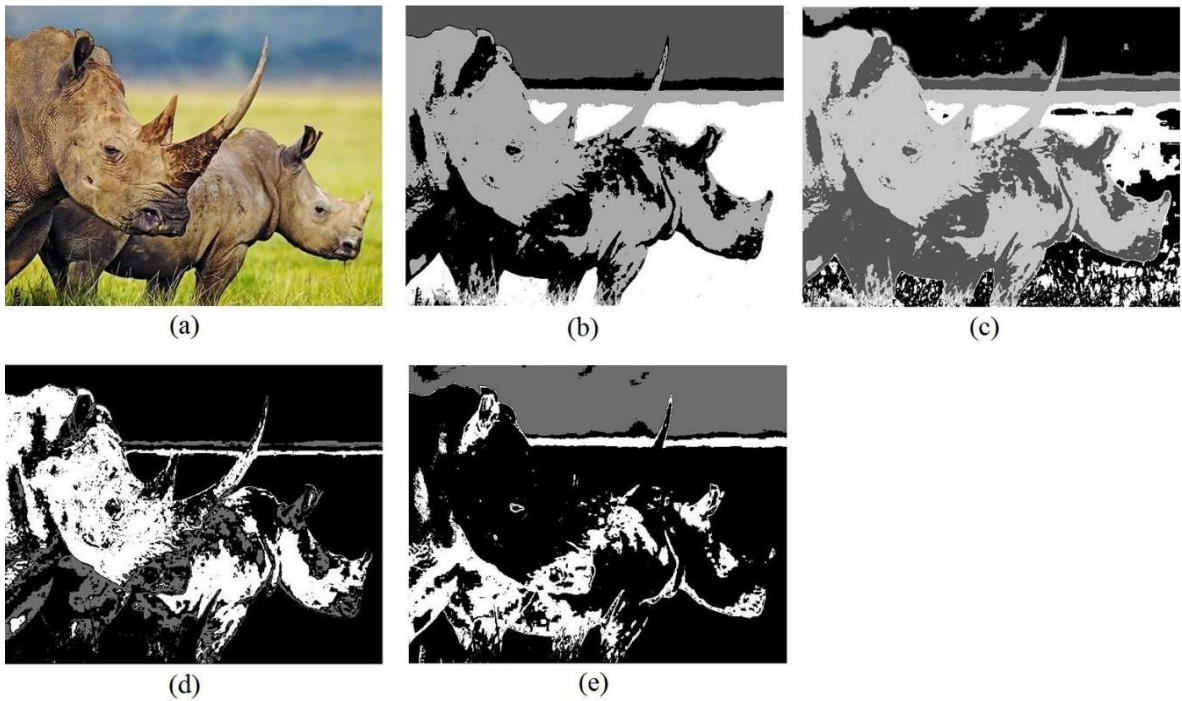

**Figure 10.** Image segmentation by FCM and AD-FCM: (a) the original picture, (b) FCM, (c) AD-FCM with certainty 0.4, (d) AD-FCM with certainty 0.5 and (e) AD-FCM with certainty 0.55

## 4.4 AD-FCM for eliminating the effect of added noise for privacy in location data

In mobile users' location privacy, one of the major concerns is to confuse the potential intruder so that he or she cannot distinguish user's real location. Hoh et al [33] propose K-means for determining the user's most visited locations such as home, work and hobbies. One of the methods for achieving privacy is to send 'noisy' queries such that the intruder is confused [34]. Here we use MSR GPS privacy dataset 2009 [35] and extract the center of clusters using the method proposed by Hoh et al. Then we add noisy queries to dataset and extract the center of clusters with FCM and AD-FCM and then calculate the sum of errors which is the distance between the real centers and extracted centers. Table 10 shows the results of this experiment and comparison of both algorithms. The results are computed by sum of error for five different users after adding noise and extracting users' important locations computed by FCM and AD-FCM with different certainty factors.

Table 10. Comparing FCM and AD-FCM for location privacy data

| User | Number of Queries (records) | Mean error for Basic FCM | Mean error for AD-FCM With certainty=0.4 | Mean error for AD-FCM With certainty=0.5 | Mean error for AD-FCM With certainty=0.6 |
|---|---|---|---|---|---|
| #2 | 122,528 | 0.3585 | 0.2615 | 0.2243 | 0.2434 |
| #3 | 418,983 | 0.6181 | 0.4773 | 0.3467 | 0.8261 |
| #4 | 148,510 | 14.4864 | 6.8033 | 5.4082 | 7.4821 |
| #5 | 391,753 | 7.5178 | 6.1961 | 5.2709 | 4.8352 |
| #6 | 31,013 | 0.7985 | 0.8048 | 0.6430 | 0.5977 |

As we expected, in AD-FCM the sum of errors is lower than of the FCM mostly between 30 to 50 percent. The variance of errors in AD-FCM for different executions are shown in Table 11. A remarkable observation here is that for certainty threshold 0.6, the variance is significantly higher than other parts. As shown in Table 11, a certainty threshold value of 0.6 did not prosper the results, and it leads to weaker results due to the fact that too many records are marked as ambiguous with this value of certainty threshold. Many of real records eliminate either, and in some cases this may cause the results to become worse or even unpredictable.

Table 11. Variance of the error for different executions of the AD-FCM for location privacy data

| User | Variance for AD-FCM With certainty=0.4 | Variance for AD-FCM With certainty=0.5 | Variance for AD-FCM With certainty=0.6 |
|---|---|---|---|
| #2 | 0.2303 | 0.0441 | 0.2764 |
| #3 | 0.0212 | 0.0177 | 0.024 |
| #4 | 0.12 | 0.8103 | 40.2326 |
| #5 | 0.0453 | 4.49 | 7.1332 |
| #6 | 0.000447 | 0.0496 | 0.1307 |

# 5 Conclusion

In the current study, we have shown that the accuracy of fuzzy clustering is improved by focusing on the concept of ambiguous data. We proposed a method for separating ambiguous output records, which cannot be clustered with high certainty by existing methods. By this approach we prosper the algorithm's accuracy. We evaluated this method on four well-known datasets in four different fields of network intrusion detection, medical diagnosis, image segmentation and added noise for privacy in location dataset. The results show that with separating ambiguous records, the false error rate decreases significantly: 41% decrease of error rate (from 34 to 20) for Pima dataset and 81.57% decrease of error rate (from 3.8 to 0.7) for IDS KDD99 and up to 50% for noisy queries added to MSR. It was also shown in image segmentation that more details of photos are explored by marking ambiguous parts, which can be used as a good basis for improving methods in this field of research. The number of clusters as well as the optimal values of certainty threshold were discussed which are based on the problem and context, but a value of about 0.4 was good for most cases. Optimal determination of the certainty threshold for different application areas remains an open problem. Future work may also include applying this method to other domains and studying the ambiguity in other fuzzy learning methods or using AD-FCM hybrid with other learning algorithms.